\documentclass[a4paper,11pt]{article}

\usepackage{fourier}
\usepackage{color}
 \usepackage{graphicx}
\usepackage{url}
\usepackage[affil-it]{authblk}
\usepackage{amsmath}
\usepackage{wrapfig}
\usepackage{bbm, dsfont}
\usepackage{subcaption}
\usepackage{amssymb}
\usepackage{booktabs}
\usepackage{adjustbox}
\usepackage[T1]{fontenc}
\usepackage{times}
\usepackage[breaklinks,colorlinks]{hyperref}
\hypersetup{
    linkcolor=black,       
    citecolor=black,     
    filecolor=black,   
}
\usepackage{authblk}
\begin{document}

\title{Self-Supervised Learning Based Handwriting Verification}

\author{%
    Mihir Chauhan\textsuperscript{††} \quad Mohammad Abuzar Hashemi\textsuperscript{¶} \quad Abhishek Satbhai\textsuperscript{§} \quad Mir Basheer Ali\textsuperscript{¶¶} \quad
    
    Bina Ramamurthy\textsuperscript{‡} \quad Mingchen Gao\textsuperscript{||} \quad Siwei Lyu\textsuperscript{†} \quad Sargur Srihari\textsuperscript{‡‡}\\ 
    
    Department of Computer Science and Engineering \\ 
    The State University of New York, Buffalo, NY, USA \\ 
    {\tt\small {\{mihirhem\textsuperscript{††}, bina\textsuperscript{‡}, mgao8\textsuperscript{||}, siweilyu\textsuperscript{†}, srihari\textsuperscript{‡‡}\}}@buffalo.edu} \\
    {\tt\small {\{ma.hashemi.786\textsuperscript{¶}, abhishek07satbhai\textsuperscript{§}, alimirbasheer\textsuperscript{¶¶}\}}@gmail.com}
}

\date{}
\maketitle
\thispagestyle{empty}

\begin{abstract}
We present SSL-HV: \textbf{S}elf-\textbf{S}upervised \textbf{L}earning approaches for the task of \textbf{H}andwriting \textbf{V}erification. This task involves determining whether a given pair of handwritten images originate from the same or different writer distribution. We have compared the performance of four generative and eight contrastive SSL approaches against handcrafted feature extractors and supervised approaches on CEDAR AND dataset. We show that ResNet based Variational Auto-Encoder (VAE) outperforms other generative approaches achieving 76.3\% accuracy, while ResNet-18 fine-tuned using Variance-Invariance-Covariance Regularization (VICReg) outperforms other contrastive approaches achieving 78\% accuracy. Using a pre-trained VAE and VICReg for the downstream task of writer verification we observed a relative improvement in accuracy of 6.7\% and 9\% over ResNet-18 supervised baseline with 10\% writer labels. Our code is publicly available at: \url{https://github.com/Mihir2/ssl-hv}
\end{abstract}


\section{Introduction}
\label{sec:intro}

Handwriting Verification is the process of comparing questioned writing and known writing. It is a critical task in various domains including forensics, banking, and legal proceedings. Traditional approaches to handwriting verification \cite{Individuality:1} includes using global handwriting features to determine between-writer and within-writer variations. These approaches relied on rule based \cite{GSCfirst:5} and handcrafted features \cite{1467360} which did not capture the full complexity and variability in handwriting. With the advent of artificial neural networks, deep networks using Convolutional Neural Networks (CNNs) \cite{Resnet:25} and Vision Transformers (ViT) \cite{dosovitskiy2021image} are now used to generate image representations. Such networks have shown promising results on variety of vision tasks and there has been increasing amount of research being done to apply these deep learning techniques to the downstream tasks of Handwriting Identification, Writer-Retrieval and Recognition. For example, supervised approaches $f_{\theta_{s}}(x, y)$ \cite{8563247} \cite{8583759} \cite{chauhan2019explanation} for handwriting verification uses deep networks heavily reliant on supervised writer labels $y$ during it's training process. But, collecting a diverse dataset $X(x_{q},x_{k}, y)$ with known $x_{k}$, questioned $x_{q}$ handwritten samples and corresponding writer labels $y$ is expensive and time-consuming. The dependency on labeled dataset limits the scalability of the supervised methods because of data collection and labeling efforts. 

\begin{figure}[t]
  \centering
   \includegraphics[width=1\linewidth]{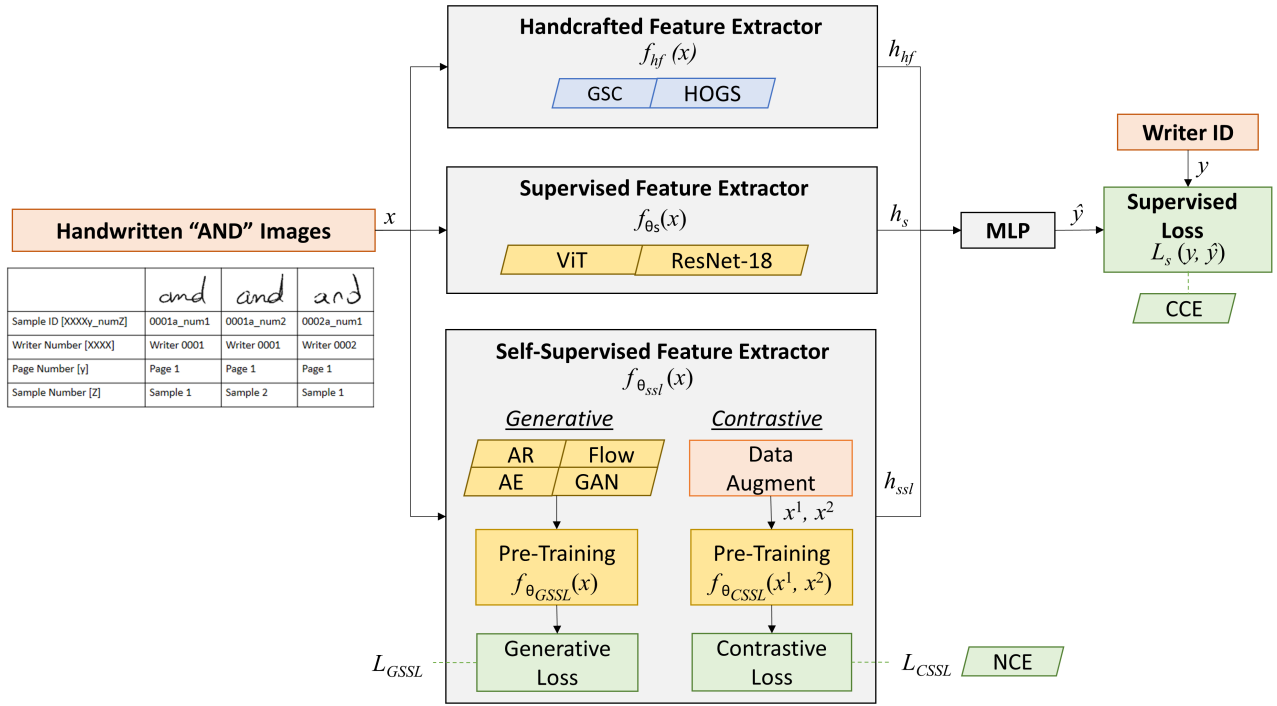}
   \caption{The overall framework for Self-Supervised based Handwriting Verification (SSL-HV)}
   \label{fig:ssl_hv}
\end{figure}

Self-Supervised Learning (SSL) provides an approach to learn meaningful representations from input $X$ by leveraging the intrinsic patterns and structures from input $X$ without the need of explicit supervised labels $y$. This helps to reduce the burden of data collection and allows the utilization of large amount of untapped unlabeled or partially labeled data that is available. Although, SSL has been employed in various domains within Computer Vision, but the application of SSL has been limited within handwriting domain. Motivated by the lack of SSL application to the domain of handwritten document representation, we apply SSL approaches $f_{\theta_{SSL}}(x)$ to generate representations $h_{ssl}$ for the handwritten images $(x_{q}, x_{k})$ as shown in Figure \ref{fig:ssl_hv}. Our contributions are as follows: 

\begin{enumerate}
    \item We create a baseline for handwriting verification task using handcrafted features (GSC, HOGS) and supervised learning approach with ResNet-18 \cite{Resnet:25} and ViT \cite{dosovitskiy2021image}.
    \item We pre-train using four Generative SSL (GSSL-HV) approaches for learning representations $h_{ssl}$ for the downstream task of Handwriting Verification using Auto-Regressive Image Modeling, Flow based model, Masked AutoEncoder, ResNet based Variational AutoEncoders (VAE) and Bi-Directional Generative Adversarial Network (Bi-GAN).
    \item We pre-train eight Contrastive SSL (CSSL-HV) approaches using ResNet-18 \cite{Resnet:25} as encoder networks $f_{\theta}(x)$ to learn representations from handwritten images $x$.
    \item Lastly, we fine-tune a MLP $f_{\theta_{MLP}}(h_{SSL},y)$ for the downstream task of handwriting verification on CEDAR AND dataset.
\end{enumerate} 
\vspace{-1.5em}

\section{Related Work}

An example of handwritten features generated using SSL is SURDS \cite{chattopadhyay2022surds}. SURDS propose a two-staged SSL framework for writer independent Offline Signature Verification by using a dual triplet loss base fine tuning. Furthermore, SSL is used in \cite{LASTILLA2022102875} for handwriting identification for medieval documents by finetuning a ResNet18 \cite{Resnet:25} architecture on a set of unlabeled manuscripts using Triplet Loss. Another example of SSL based pre-training is POSM \cite{mehralian2023selfsupervised} handwritten representations are extracted from online handwriting in English and Chinese languages. The pretrained POSM models were capable of achieving good results on diverse set of handwriting tasks such as writer identification, handedness classification etc. The authors in \cite{inbook} use SSL approach with Vision Transformers (ViT) for writer retrieval task based on knowledge distillation. The authors also showcased the attention feature maps which elaborated different parts of the handwriting like loops, characters which enhanced the explainability for writer retrieval. More recently, CSSL-RHA \cite{wang2023csslrha} uses Contrastive SSL for Handwriting Authentication wherein a hybrid CNN-Vit based network is used for pre-training with momentum-based paradigm followed by projection head to minimize a variant of NCE \cite{10.5555/2188385.2188396} loss. 

\section{Dataset}
CEDAR AND dataset is used for pre-training and fine-tuning on a downstream task of handwriting verification. CEDAR AND dataset is derived from CEDAR Letter dataset wherein 1567 writers have written a letter manuscript three times. Each manuscript had up to five occurrences of the word ``AND''. The manuscripts were passed to transcript-mapping tool of CEDAR-FOX to extract image fragments of the word ``AND''. In total, the tool was able to extract 15,518 image fragments of the word ``AND''. Some examples of the snippets are shown in Figure \ref{fig:ssl_hv}. Each image was resized to 64x64 keeping uniform padding and aspect ratio. 

Handcrafted features (GSC, HOGS) were derived for each ``AND'' image fragment. Gradient Structural Concavity (GSC) \cite{GSCfirst:5} features are used as micro features for forensic verification by CEDAR-FOX tool, a state-of-the-art handwriting analysis tool \cite{Srihari_Srinivasan_Desai_2018}, developed at Center of Excellence for Document Analysis and Recognition, University at Buffalo. GSC features were extracted for the binarized ``AND'' images using a C-code. The GSC features are in 512 dimensions. CEDAR-FOX uses the GSC features to verify the log likelihood ratio (LLR) between the known and questioned handwritten sample. We  use OpenCV HOGDescriptor for generating Histogram of Oriented Gradients (HOGS) \cite{1467360} features for each AND image fragment. The HOGS features are 1764 dimensional vectors.

We use unseen writer data partitioning for supervised fine-tuning for downstream verification task. Hence, $W_{train} \bigcap W_{test} = \emptyset$, where $W_{train}$ represents train writers and $W_{test}$ represents test writers. For both pre-training and downstream fine-tuning, writer ids $w_{i}$ up-to 1200 were used for training $i_{train} \in \{{1,2,..,1200}\}$ and rest were used as test. For fine-tuning, we generate equal number of same and different writer sample pairs. We have two setups for fine-tuning with 10\% and 100\% of train writers resulting in 13,232 and 129,602 pairs of known and questioned ``AND'' samples. The test is fixed for both setups with all the test writers $i_{test} > 1200$.    

\section{Supervised Learning for Handwriting Verification (SL-HV)}
Supervised baseline is performed using two handcrafted features (GSC, HOGS), CNN based ResNet-18 \cite{Resnet:25} and Vision Transformer based architecture on CEDAR AND dataset for 10\% and 100\% of train writers. ResNet-18 CNN architecture having 11.2 M parameters is fine-tuned on the training pairs for each setup. We update the first convolutional layer to accept 3 channel input and set the last FC within ResNet-18 to be Identity and add a supervised classification head as elaborated in the next section. MaskedCausalVisionTransformer is used as the ViT architecture having 88.2M parameters. The output of the feature extractors is fed into 2 fully-connected (FC) layers. FC1 and FC2 has 256 and 128 hidden neurons with ReLU activations. The final layer has 2 output neurons whose softmax activations represent similarity of samples with a one hot vector representation. We use categorical cross entropy loss given one-hot encoded logits compared to the target which is binary (0 or 1). Batch Size for training was 256, Learning rate 1e-3, Adam Optimizer and Early stopping with F1 score stagnating with patience 5 and delta 0.001 for all baselines GSC, ViT and ResNet-18. The results from the supervised training with on 10\% and 100\% train writers are tabulated in Table \ref{table:supervised_baseline}. We compare the downstream model performance using classification metrics to compare pre-trained SSL-HV model performance against handcrafted features and supervised models. 

\begin{table}[ht!]
\centering
\begin{adjustbox}{width=0.6\textwidth}
\small
    \begin{tabular}{l|c|c|c|c}
    \toprule \toprule
    \textbf{Model} & \textbf{Accuracy} & \textbf{Precision} & \textbf{Recall} & \textbf{F1-Score} \\ \midrule
    GSC \cite{GSCfirst:5}      & 0.71 / 0.78	& 0.69 / 0.81	& 0.72 / 0.77	& 0.69 / 0.79  \\
    ResNet-18 \cite{Resnet:25} & 0.72 / 0.84	& 0.70 / 0.86	& 0.73 / 0.82	& 0.72 / 0.84  \\
    ViT  \cite{dosovitskiy2021image}      & 0.65 / 0.79	& 0.68 / 0.80   & 0.64 / 0.78   & 0.66 / 0.79  \\ \bottomrule \bottomrule
    \end{tabular}
\end{adjustbox}
\caption{Performance Metrics on Test Writer set for Supervised Baselines with 10\% \& 100\% of Train Writers.}
\label{table:supervised_baseline}
\end{table}

\section{Learning Representation using Self-Supervised Learning based Pre-Training}

SSL has been used as a pre-text task for pre-training a network $f_{\theta_{SSL}}$ to generate representations $h_{ssl}$ only using input data $x$ without explicit writer labels $y$. SSL approaches are classified into Generative and Contrastive: 

\vspace{-1em}

\subsection{Generative SSL for Handwriting Verification (GSSL-HV)}

Generative SSL approaches pre-trains a network $f_{\theta}(x)$ which learns to generate hidden latent representation $h$ from underlying probability distribution $P(X)$ by maximizing likelihood of generative objective function $L$. We have compared four generative SSL approaches: Auto-Regressive (AR) models, Flow-based models, Auto-Encoding models, and Generative Adversarial Networks (GANs).  

\textbf{AR Models} \cite{elnouby2024scalable} are directed probabilistic graphical models which models input data distribution $p(X) = p(x_{1}, ..., x_{n})$ as product of conditionals $\prod_{i=1}^{n} P(x_{i}|(x_{1}, .., x_{i-1}))$ where $n$ is the input dimensionality.
We have used state-of-the-art AR model Auto Regressive Image Modeling (\textbf{AIM}) \cite{elnouby2024scalable}. AIM uses AutoRegressive loss function $L_{AR}$ using ViT architecture. AIM uses random masking to sequentially learn masked patches of the input image with non-overlapping image patches of size 32. We resize input to 224x224 and normalize the images for training and testing. We use CausalVisionTransformer as the backbone with masked causal attention with sequence length as 49 with no [CLS] classification token. The embedding dimension of the encoder is 768, depth and number of attention heads 12.

\textbf{Normalizing Flow} \cite{ho2019flow} learns the true data distribution $p(x)$ by a sequence of invertible transformation functions $f(x)$ to map input $x$ to latent representation $z$ by minimizing the negative log-likelihood over the input distribution dataset $p(x)$. The input handwritten AND image is inverted and converted to gray-scale with a single channel and pixel value between [0,255]. We use a single variational de-quantization layer to quantize discrete pixel values as samples from continuous distribution which helps improve diversity and quality of generated samples. This is followed by 2 affine coupling layers with a single channel checkerboard mask throughout the network. A Gated CNN using two-layer convolutional ResNet block with input gate. We use multi-scale architecture using Squeeze and Split layers. The model is evaluated based using bits per dimension (bpd) for the train and validation set. Train and Val bpd was 0.836 and 0.841 respectively.

\textbf{MAE} Masked AutoEncoder \cite{he2021masked} applies random mask patches $M$ with high masking ratio on the input image $x$. Encoder $f_{enc}(x^{m}, p)$ is a ViT which takes as input visible parts of the image $x^{m}$ and positional embeddings $p$ to generates latent representation $z^{m}$. A lightweight decoder $f_{dec}(z^{m}, p, M)$ reconstructs the entire image $x$ including the missing patches using the encoded latent representations $z^{m}$ of the masked image $x^{m}$, positional embeddings $p$ and tokens from masked patches $M$. The loss is computed using Mean Squared Error (MSE). MAE is applied using random masking with masking ratio set to 20\%. We use vit-base-patch32-224-in21k \cite{wu2020visual} as the backbone for MAE with 768 embedding dimension. \textbf{VAE} Variational Auto Encoder \cite{Kingma2014AutoEncodingVB} maximizes $p(x)$ by sampling from latent dimensions $z$ originating from posterior probability $p(z|x)$ while minimizing the Reconstruction and Latent loss. VAE takes as input an inverted image with size 64x64x3. The encoder $f_{enc}$ and decoder $f_{dec}$ used in VAE is the ResNet Encoder and Decoder by Pytorch Bolt \cite{jirka_borovec_2022_7447212}. $f_{enc}$ is followed by two $FC_{\mu}$ and $FC_{\sigma}$. $f_{enc}$ output dimensions are 512 and latent dimensions $z$ are 256.

\textbf{BiGan} Bi-Directional GAN \cite{donahue2017adversarial} is trained using an Encoder $f_{enc}$, Generator $f_{G}$ and Discriminator $f_{D}$ network to learn representation from handwritten images. The $f_{enc}$ network consists of 5 blocks, starting with 1024 hidden units and consequently layers have hidden units divided by 2. $f_{enc}$ network which takes as input raw flattened image and outputs latent representation $z$ with dimensionality as 100. The $f_{G}$ network takes as input $z$ latent representation and also network structure which is opposite to the $f_{enc}$ in order to re-generate back the output dimensions 64x64 with 3 channels. The discriminator takes as input image and latent representation and minimizes the binary cross-entropy loss between the fake and valid combination samples of latent representation and input image. We use pytorch lightning multi-optimizer function to update the gradients of generator and discriminator in an alternating fashion. 

\begin{figure}[t]
  \centering
  \begin{subfigure}{0.1\linewidth}
    \centering
    \includegraphics[width=0.7\linewidth]{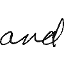}
    \caption{Original}
  \end{subfigure}%
  \begin{subfigure}{0.1\linewidth}
    \centering
    \includegraphics[width=0.7\linewidth]{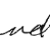}
    \caption{Crop+R}
  \end{subfigure}%
  \begin{subfigure}{0.1\linewidth}
    \centering
    \includegraphics[width=0.7\linewidth]{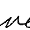}
    \caption{Center}
  \end{subfigure}%
  \begin{subfigure}{0.1\linewidth}
    \centering
    \includegraphics[width=0.7\linewidth]{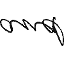}
    \caption{V. Flip}
  \end{subfigure}%
  \begin{subfigure}{0.1\linewidth}
    \centering
    \includegraphics[width=0.7\linewidth]{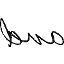}
    \caption{H. Flip}
  \end{subfigure}\\[1ex]
  \begin{subfigure}{0.1\linewidth}
    \centering
    \includegraphics[width=0.7\linewidth]{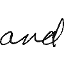}
    \caption{Rotate}
  \end{subfigure}%
  \begin{subfigure}{0.1\linewidth}
    \centering
    \includegraphics[width=0.7\linewidth]{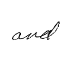}
    \caption{Persp.}
  \end{subfigure}%
  \begin{subfigure}{0.1\linewidth}
    \centering
    \includegraphics[width=0.7\linewidth]{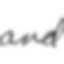}
    \caption{Blur}
  \end{subfigure}%
  \begin{subfigure}{0.1\linewidth}
    \centering
    \includegraphics[width=0.7\linewidth]{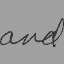}
    \caption{Jitter}
  \end{subfigure}%
  \begin{subfigure}{0.1\linewidth}
    \centering
    \includegraphics[width=0.7\linewidth]{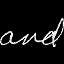}
    \caption{Invert}
  \end{subfigure}
  \caption{Data Augmentation Views from an example original image of word ``AND".}
  \label{fig:data_aug_viz}
\end{figure}

\vspace{-1.0em}

\subsection{Contrastive SSL for Handwriting Verification (CSSL-HV)}
Contrastive learning uses discriminative approach to learn representations $h$ by maximizing the agreement between similar (positive) images and minimize the agreement between dissimilar (negative) images $P(Y|X=x)$. These discriminative model use Noise Contrastive Estimation (NCE) \cite{10.5555/2188385.2188396} loss whose aim is to compare and learn the objective function as shown in Eqn. \ref{eq:nce_loss} below:
\begin{equation}
\mathcal{L}_{nce}=-\log \frac{\exp \left[h_{a}^T \cdot h_{+} / \tau\right]}{\exp \left[h_{a}^T \cdot h_{+} / \tau\right]+ \exp \left[h_{a}^T \cdot h_{-} / \tau\right]} 
\label{eq:nce_loss}
\end{equation}

where, $h = f(x)$ are features and $f$ is a function to embeding input $x$. $x^{+}$ is similar to input image used an an anchor $x^{a}$, $x^{-}$ is dissimilar to $x$ and $f$ is a function to embed input image $x$ to features $h$. 

\textbf{Data Augmentation} regularizes the model and helps to learn patterns within different parts of the input image. Figure \ref{fig:data_aug_viz} shows a variety of augmentation techniques which can be used to get different forms of invariances for the ``AND'' image dataset. We choose the data-augmentations for the domain of handwriting verification such that the writer characteristics are not perturbed (e.g. scale). We have used Pytorch Transforms \cite{pytorch_transform} implementation for data augmentation.

\textbf{Pre-Training, Projection Network and Loss Function} For CSSL pre-training, we use a ResNet-18 \cite{Resnet:25} network with stochastic gradient descent using custom loss functions. For CEDAR ``AND'' dataset we chose a larger kernel size 7x7 variant of ResNet-18 since the Crop size from augmentation is 224x224. The variant also has 11.2M parameters. For the projection head/network and loss function, each CSSL approach described will have different number of neurons and layers in the projection network. We have experimented with eight CSSL methods: MoCo\cite{he2020momentum}, SimCLR\cite{chen2020simple}, SimSiam \cite{chen2020exploring}, FastSiam \cite{fastsiam2022}, DINO\cite{caron2021emerging}, BarlowTwins \cite{zbontar2021barlow} and VicReg\cite{bardes2022vicreg}. All the experiments were conducted using Lightly SSL Python package \cite{susmelj2020lightly} on AWS notebook instance with ml.g5.2xlarge which has 1 Nvidia A10G (24GB) GPU. More details about the exact data augmentations used within each CSSL-HV approach is available on github.

\vspace{-0.8em}

\subsection{Pre-Training Metric for GSSL-HV and CSSL-HV} 
We use a separation metric defined by the difference between mean $\text{COS}_{inter}$ and $\text{COS}_{intra}$ writer similarity to track SSL model capability to differentiate samples between writers. 
\begin{equation}
    \text{COS}(\mathbf{h}_k, \mathbf{h}_q) = \frac{\mathbf{h}_k \cdot \mathbf{h}_q}{\|\mathbf{h}_k\| \|\mathbf{h}_q\|}
\label{eq:intra_inter_cosine_similarity}
\end{equation}
In the Equation \ref{eq:intra_inter_cosine_similarity} above,  $\mathbf{h}_q$ represents the features of the handwritten sample from known writer and $\mathbf{h}_q$ represents the features of the handwritten samples from questioned writer. Table \ref{table:ssl_results} shows $\text{Intra-Nd}$ and $\text{Inter-Nd}$ cosine similarities calculated on all embedding (Nd). The table also shows the $\text{Intra-2d}$ and $\text{Inter-2d}$ cosine similarity calculated on 2-dimensions with the help of TSNE dimensionality reduction.

\begin{table}[t]
\centering
\begin{adjustbox}{width=0.7\textwidth}
    \begin{tabular}{@{}l|c|c|c|c|c}
    \toprule \toprule
    \textbf{Model} & \textbf{Intra-Nd} & \textbf{Inter-Nd} & \textbf{Intra-2d} & \textbf{Inter-2d} & \textbf{Accuracy} \\ \midrule
    Raw Pixels                           & 0.96   & 0.95  & 0.07  & -0.02  & 0.63  \\
    HOGS \cite{1467360}                  & 0.57   & 0.02  & 0.63  & 0.11   & 0.72  \\
    GSC \cite{GSCfirst:5}                & 0.92   & 0.67  & 0.86  & 0.56   & 0.71  \\\midrule
    AIM \cite{elnouby2024scalable}       & 0.32   & -0.05 & 0.78 & 0.75    & 0.73  \\  
    Flow \cite{ho2019flow}  & 0.12 & 0.08  &  0.12 & 0.01  & 0.66 \\ 
    MAE \cite{he2021masked}   & 0.18  & 0.02  & 0.82  & 0.77  & 0.71 \\
    \textbf{VAE} \cite{Kingma2014AutoEncodingVB}   & \textbf{0.24}  & \textbf{0.06}  & \textbf{0.38} &  \textbf{0.30} &  \textbf{0.75}  \\ 
    BiGAN \cite{donahue2017adversarial} & 0.35  & 0.30  &  0.27 & 0.25 & 0.68 \\ \midrule
    MoCo \cite{he2020momentum}           & 0.89   & 0.78  & 0.92  & 0.73   & 0.73  \\
    SimClr \cite{chen2020simple}         & 0.89   & 0.87  & 0.87  & 0.85   & 0.72  \\
    BYOL \cite{grill2020bootstrap}       & 0.88   & 0.84  & 0.91  & 0.97   & 0.73 \\
    SimSiam \cite{chen2020exploring}     & 0.87   & 0.81  & 0.94  & 0.84   & 0.75 \\
    FastSiam \cite{fastsiam2022}         & 0.83   & 0.75  & 0.83  & 0.75   & 0.71 \\
    DINO \cite{caron2021emerging}        & 0.88   & 0.85  & 0.78  & 0.74   & 0.68 \\
    BarlowTwins \cite{zbontar2021barlow} & 0.87   & 0.79  & 0.66  & 0.38   & 0.76 \\
    \textbf{VicReg} \cite{bardes2022vicreg}  & \textbf{0.69}   & \textbf{0.48}  & \textbf{0.65}  & \textbf{0.60}   & \textbf{0.78}\\
    \bottomrule \bottomrule
    \end{tabular}
\end{adjustbox}
\caption{Performance comparison of GSSL-HV and CSSL-HV approaches against handcrafted feature baselines on CEDAR AND Dataset with 10\% train writers.}
\label{table:ssl_results}
\end{table}

\vspace{-1em}

\section{Results}
\vspace{-0.7em}
Table \ref{table:ssl_results} shows the performance of the SSL and baseline approaches. We observe that higher the separation between writers (Intra distance - Inter distance) leads to higher test accuracy on a small training dataset. In the experiments performed we observe VAE to be best performing with a good separation of 0.18 between the intra-inter distance between and amongst the writers during the pre-training phase which lead to 6.7\% relative increase in the accuracy when compared to the best performing supervised ResNet-18 baseline with accuracy 72\% accuracy on 10\% train writers. AIM and MAE outperformed it's supervised counterpart ViT on 10\% train writers but had a lower precision compared to VAE which is contributed to the difference in the feature extraction process. Within VAE we used ResNet-18 architecture whereas AIM and MAE uses a ViT architecture whose baseline metrics under performed when compared to ResNet-18 as shown in Table \ref{table:supervised_baseline}. Flow based models performed similar to the baselines but under-performed when compared to VAE and AIM. This is contributed to the fact that flow based models do not support sparsity in feature representation and the type of invertible transformations are not suitable for granular variations within handwritten styles. GANs also performed similar to baselines, this may be primarily due to the fact that GANs are primarily used for data generation and do not naturally include an encoder to map data back to the latent space. We observed that GSC and HOGS features have maximum separation between $\text{COS}_{intra}$ and $\text{COS}_{inter}$ whereas using raw pixels the separability is very low. From table \ref{table:ssl_results} maximum separation of 0.28 is obtained using VicReg on the CEDAR AND Dataset leading to 9\% relative improvement in accuracy over best performing supervised ResNet-18 baseline. 

\vspace{-1em}

\section{Conclusion}
\vspace{-0.7em}
SSL provides a pathway to generating robust handwritten features which helps improve downstream task of handwriting verification with limited amount of training labels. In this paper, we evaluated AutoRegressive, Flow Based, AutoEncoding and GANs as part of the GSSL-HV framework. We also compared performance of eight CSSL-HV approaches. VAE outperformed other generative self-supervised feature extraction approaches, achieving a relative gain of 6.73\% in accuracy whereas VICReg was outperformed all the generative and contrastive approaches with a relative accuracy gain of 9\% over the baselines. Future research can aim to enhance the feature extraction capabilities using full letter manuscripts and multiple unlabeled handwritten datasets such as IAM handwriting dataset for comparing similar and different handwritten content using SSL approaches.



\bibliographystyle{apalike}
\bibliography{references}

\end{document}